\documentclass[letterpaper, 10 pt, conference]{ieeeconf} 
\usepackage{times}
\usepackage{comment}
\usepackage{graphicx}
\usepackage{amsmath}
\usepackage{amssymb}
\usepackage{graphicx}
\usepackage{url}
\usepackage{array}
\usepackage{multirow}
\usepackage{color}
\usepackage{fancyhdr}
\usepackage{threeparttable}

\newcommand{\etal}{\textit{et al}. }
\usepackage{amsmath}
\usepackage[mathscr]{eucal}

\usepackage{multicol}
\usepackage[noend]{algpseudocode}
\usepackage{algorithm}
\usepackage{subfigure}

\IEEEoverridecommandlockouts                              % This command is only
\title{An Information-rich Sampling Technique over Spatio-Temporal CNN for Classification of Human Actions in Videos
}

\author{S H Shabbeer Basha, Viswanath Pulabaigari, Snehasis Mukherjee \\
Indian Institute of Information Technology Sri City, India% <-this % stops a space
\thanks{S.H.S. Basha, Viswanath P., S. Mukherjee  are with Computer Vision and Machine Learning Groups, Indian Institute of Information Technology, Sri City, Andhra Pradesh - 517646, India. 
{\tt\small email: \{shabbeer.sh, viswanath.p, snehasis.mukherjee\}@iiits.in}}%
}

\begin{document}

\maketitle

% \thispagestyle{specialfooter}
% \pagestyle{empty}
% \lipsum[1-30]

%%%%%%%%%%%%%%%%%%%%%%%%%%%%%%%%%%%%%%%%%%%%%%%%%%%%%%%%%%%%%%%%%%%%%%%%%%%%%%%%
\begin{abstract}
We propose a novel scheme for human action recognition in videos, using a 3-dimensional Convolutional Neural Network (3D CNN) based classifier. Traditionally in deep learning based human activity recognition approaches, either a few random frames or every $k^{th}$ frame of the video is considered for training the 3D CNN, where $k$ is a small positive integer, like $4$, $5$, or $6$. This kind of sampling reduces the volume of the input data, which speeds-up training of the network and also avoids over-fitting to some extent, thus enhancing the performance of the 3D CNN model. In the proposed video sampling technique, consecutive $k$ frames of a video are aggregated into a single frame by computing a Gaussian-weighted summation of the $k$ frames. The resulting frame (aggregated frame) preserves the information in a better way than the conventional approaches and experimentally shown to perform better. In this paper, a 3D CNN architecture is proposed to extract the spatio-temporal features and follows Long Short-Term Memory (LSTM) to recognize the human actions. The proposed 3D CNN architecture is capable of handling the videos where the camera is placed at a distance from the performer. Experiments are performed with KTH and WEIZMANN human actions datasets, whereby it is shown to produce comparable results with the state-of-the-art techniques.

% The invention of Convolutional Neural Networks (CNN) makes a landmark in the area of Computer Vision to solve a wide variety of image classification problems, including object recognition, localization, segmentation, human action recognition, and many more. However, there are still many gaps specific to image classification problem remained to be unsolved. In this paper, we address the following gaps in the state-of-the-art CNNs, impacting the Fully Connected (FC) layers of CNN architecture in image classification: (i) \textit{How the deeper/wider datasets influence the necessity of FC layers in CNN?}, (ii) \textit{What is the impact of deeper/shallow architectures on the number of required FC layers?}, and (iii) \textit{Which kind of architecture (deeper/ shallower) is better suitable for which kind if (deeper/ shallower) datasets}. To address the above gaps in the existing CNN, we have performed extensive experiments with three popular CNN architectures having different depths. Four widely adapted datasets including CIFAR-10, CIFAR-100, Tiny ImageNet, and CRCHistoPhenoTypes are used to justify the impact of fully-connected (FC) layers for image classification. 
\end{abstract}

\section{Introduction and Related Works}
\label{introduction}
%%%%%%%%%%%%%%%%%%%%%%%%%%%%%%%%%%%%%%%%%%%%%%%%%%%%%%%%%%%%%%%%%%%%%%%%%%%%%%%%
Human action recognition in videos has been an active area of research, gaining the attention of Computer Vision and Machine Learning researchers during the last decade due to its potential applications in various domains, including intelligent video surveillance systems, {\em viz.,} Human-Computer Interaction (HCI), robotics, elderly and child monitoring systems and several other real-world applications. However, recognizing human actions in the real world remains a challenging task due to several challenges involved in real-life videos, including cluttered backgrounds, viewpoint variations, occlusions, varying lighting conditions and many more. 
% The recognition task becomes even complex when the video is captured at a distance from the camera. 
This paper proposes a technique for human activity recognition in videos, where the videos are captured by a camera placed at a distance from the performer.

The approaches for recognizing human actions from videos, found in the literature, can be broadly classified into two categories~\cite{ziaeefard2015semantic}. The first, make use of motion-related features (low, mid, and high level) for human action recognition \cite{laptev2007retrieving,dollar2005behavior}. The other set of approaches experiment to learn a proper representation of the spatio-temporal features the during action using deep neural networks \cite{Sun_2017_ICCV,simonyan2014two,wang2016temporal,baccouche2011sequential}.

Handcrafted features played a key role in various approaches for activity recognition \cite{nazir2018bag}. Semantic features ease to identify similar activities that vary visually but have common semantics. Semantic features during an action contain human body parts (posture and poselet), background, motion and other features incorporating human perceptual knowledge about the activities. A study by Ziaeefard \etal \cite{ziaeefard2015semantic} examined human action recognition approaches using semantic features. Malgireddy \etal \cite{malgireddy2013language} proposed a hierarchical Bayesian model which interconnects low-level features in videos with postures, motion patterns, and categories of activities. 
% They have used a probabilistic framework for identifying and localizing the predefined activities in a video sequence, which is comparable to the use of filter models for key-word discovery in speech processing.
Very recently, Nazir \etal \cite{nazir2018bag} proposed a Bag of Expression (BOE) framework for activity recognition. 
 
The most common handcrafted feature, used for action recognition, is optical flow \cite{chaudhry2009histograms,mukherjee2015human,mukherjee2,wang2013dense}. Chaudhry \etal \cite{chaudhry2009histograms} introduced the concept of Histogram of Oriented Optical Flow (HOOF) for action recognition, where the optical flow direction is divided into octants. Mukherjee \etal \cite{mukherjee2} proposed Gradient-Weighted Optical Flow (GWOF) to limit the effect of camera shaking, where the optical flow of every frame is multiplied by the image gradient. Wang \etal \cite{wang2013dense} introduced another approach to reduce the camera shaking effect, called Warped Optical Flow (WOF), where gradient is computed on the optical flow matrix. In \cite{mukherjee2015human}, the effect of background clutter is reduced by multiplying Weighted Optical Flow (WOF) features with the image gradients. Optical flow based approaches help in dissecting the motion, but gives too much unnecessary information such as, motion information at all the background pixels, which reduces the efficacy of the action recognition system in many cases.

Spatio Temporal Interest Points (STIP) introduced by \cite{laptev2005space}, identifies spatio-temporal interest points based on the extension of Harris Corner Detection approach \cite{harris1988combined} towards the temporal domain. Several researchers have shown interest to recognize human actions with the help of some other variants of spatio-temporal features like Motion- Scale Invariant Feature Transform (MoSIFT) \cite{chen2009mosift} and sparse features \cite{dollar2005behavior}. A study on STIP based human activity recognition methods is published by Dawn \etal \cite{dawn2016comprehensive}. However, such spatio-temporal features are unable to handle the videos taken in real-world which suffers from background clutter and camera shake. Buddubariki \etal \cite{buddubariki2016event} combined the benefits of GWOF and STIP features by calculating GWOF on the STIP points. In \cite{abdulmunem2016saliency}, combination of 3-dimensional SIFT and HOOF features are used along with support vector machine (SVM) for classifying human actions.

Recently, deep learning based models are gaining the interest of researchers for recognizing human actions \cite{simonyan2014two, baccouche2011sequential,ji20133d,taylor2010convolutional,tran2015learning,karpathy2014large}. Taylor \etal \cite{taylor2010convolutional} proposed a multi-stage network, where in a Convolutional Restricted Boltzmann Machine (ConvRBM) retrieves motion-related information from each pair of successive frames at the initial layer. In \cite{simonyan2014two}, a two-stream convolutional network is proposed that comprises spatial-stream ConvNet and temporal-stream ConvNet. Ji \etal \cite{ji20133d} introduced a 3-dimensional CNN architecture for action recognition, where a 3 dimensional convolutions are used to extract the spatio-temporal features. Tran \etal \cite{tran2015learning} enhanced 3D CNN model by applying Fisher vector encoding scheme on the learned features. Karpathy \etal \cite{karpathy2014large} proposed a deep neural network for spatio-temporal resolutions: high and low resolutions, then merged them to train the CNN. 
% A study by \cite{zhu2016handcrafted} provides detailed analysis and comparison among handcrafted and deep learning based action recognition approaches. 
Kar \etal \cite{kar2017adascan} proposed a technique for temporal frame pooling in a video for human activity recognition. A survey by Herath \etal \cite{herath2017going} discusses both engineered and deep learning based human action recognition techniques.

In the literature of human action recognition, researchers have used either the fully observed video or a portion of the video to train the deep neural networks. Training the models using a portion of the video will take less amount of training time compared to training the model using entire video. However, considering a portion of the video (considering 9 frames from the entire video as in \cite{baccouche2011sequential} and 7 frames as in \cite{ji20133d}) results in information loss. Srivastava \etal \cite{srivastava2015unsupervised} used multi layer LSTM network to learn the representations of video sequences. Recently, Bilen \etal \cite{bilen2016dynamic} introduced dynamic image, a very compact representation of video used for analyzing the video with CNNs. However, dynamic images eventually dilute the importance of spatial information during action. The proposed sampling technique for video frames preserves both spatial and temporal information together.

Baccouche \etal \cite{baccouche2011sequential} proposed a completely automated deep learning architecture for KTH dataset \cite{schuldt2004recognizing}, which figures out how to characterize human activities with no earlier information. 
% However, this method considers just $9$ frames from a video to train the 3D CNN model. 
This 3D CNN architecture learns spatio-temporal features automatically, then LSTM network \cite{gers2002learning} is used to classify the learned features. Motivated by the method introduced in \cite{baccouche2011sequential}, we propose a 3D CNN to learn spatio-temporal features and then apply LSTM to classify human actions. The proposed method uses small sized filters throughout the 3D CNN architecture, which helps to learn minute information present in the videos, which can help in recognizing the action of performers appearing very small in the video, due to the distance of the camera.

Our contributions in this paper are two-folds. First, a novel sampling technique is introduced to aggregate the entire video into a fewer number of frames. Second, a 3D CNN architecture is proposed for better classification of human actions in videos where the performer looks significantly small. The choice of smaller filter size enables the proposed model work well in such scenarios where the performer looks small due to distance from the camera. We experiment with the proposed deep learning model with transfer learning technique, by transferring the knowledge learned from KTH dataset to fine-tune over WEIZMANN dataset and vice versa. 

The proposed pre-processing method is presented in Section \ref{GWF}. Section \ref{section:3dcnn} illustrates the proposed 3D CNN architecture. The experiments and results are described in Section \ref{sec:results}. Finally, Section \ref{sec:conclusion} concludes and provides scope for future research.   
%We have evaluated the proposed model on WEIZMANN dataset. As another experiment, we have used a pre-trained model of WEIZMANN dataset to fine-tune the model over KTH dataset and vice-versa. In both the cases, the proposed method outperforms the competing methods.

\section{Pre-processing using An Information Sampling Approach}
\label{GWF}

\begin{figure*}[!t]

\centering
\includegraphics[width=\textwidth,height=2.5in]{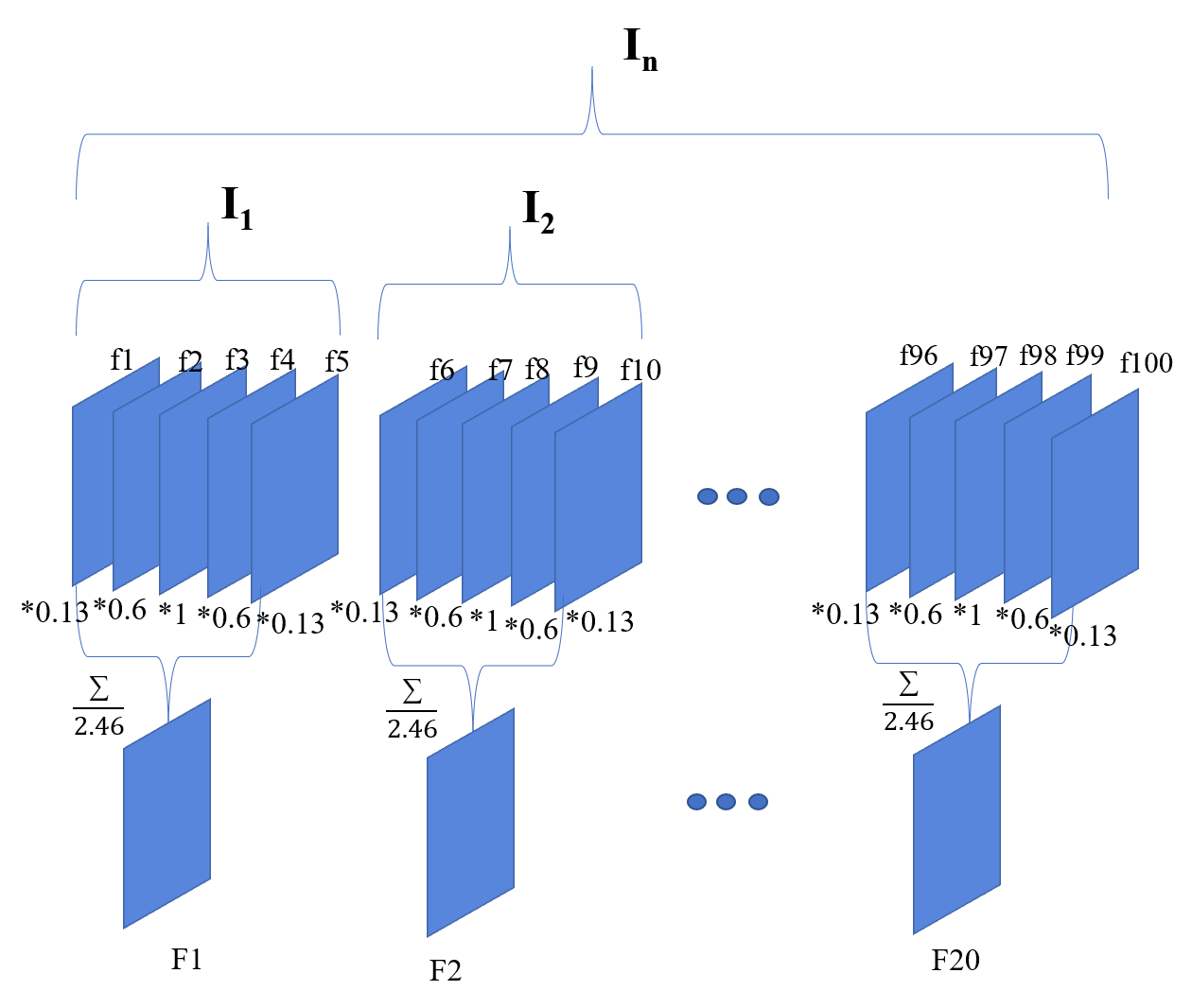}
\caption{The proposed pre-processing procedure using Gaussian Weighing Function. An entire video (collection of all frames) is represented as an exhaustive non-overlapping sequence \textbf{$I_n$}, which further has sub-sequences \{$I_1$, $I_2$, $\cdots$\}. A single pre-processed frame (for example $F1$) is obtained by performing weighted summation of consecutive five frames (for instance $f1$, $f2$, $f3$, $f4$, and $f5$ belongs to sub-sequence $I_1$) as shown in equation \ref{eq:GWF}.}
\label{GWF_fig}
\end{figure*}

The primary objective of this pre-processing step is to reduce the amount of training time and at the same time motion information should be given utmost importance. We propose a novel sampling technique to aggregate a large number of frames into a fewer set of frames using Gaussian Weighing Function (GWF), which minimizes the information loss. The proposed video pre-processing scheme is shown in Figure \ref{GWF_fig}.

Gaussian Weighing Function (GWF) is used to aggregate the entire video into a fewer number of frames. Let us consider $\{I_n\}_{n\in N}$, an exhaustive non-overlapping sequence (collection of all frames of a video), which is given by

\begin{equation}
\label{eq:subsequence}
\{I_n\} = \{I_1, I_2, \dots, I_k, \dots \},
\end{equation}
where $\{I_k\}$ is the $k^{th}$ sub-sequence of $\{I_n\}$ and $k<n$. 
Mathematically, Gaussian Weighing Function $G$, for a sub-sequence $\{I_k\}$, is given as follows:
\begin{equation}
\label{eq:GWF}
G(I_k,W) = \sum_{j=1}^{5}{I_{k{_j}}}*\frac{W_j}{\sum_{j=1}^{5}W_j}.
\end{equation}
The function $G$ takes a sub-sequence $\{I_k\}$, and Gaussian weight vector $W$ as input, and aggregates the information into a single frame. Here $W_j$ represents the $j^{th}$ element of Gaussian weight vector $W$. For example, if the size of the Gaussian weight vector (i.e.,  $\normalsize{W}$) is $5$ and the sub-sequence is $\{I_k\}$, which has five frames of the video. The vector $W$ is given by $W$ = $[0.13, 0.6, 1, 0.6, 0.13]$. A single frame is obtained by performing weighted summation of the five frames belonging to the sub-sequence $\{I_k\}$ as shown in equation (\ref{eq:GWF}). In other words, five frames are aggregated into a single frame using Gaussian weighing function. Similarly, the same process is repeated for subsequent five frames belonging to the next sub-sequence and so on. This sampling approach reduces the volume of data for training the deep learning model and also preserves the information in better way which helps to obtain better results in human activity recognition.
% , especially when the object (performer) looks small in the video.

\section{Spatio-Temporal Features Extraction using Deep Learning Models}
\label{section:3dcnn}
In this section, initially we describe 2-D CNNs, and then we present a detailed discussion about the proposed 3-D CNN architecture, which learns the spatio-temporal features.

\subsection{Convolutional Neural Networks}
There are two major problems with Artificial Neural Networks (ANN) while dealing with real world data like images, videos, and any other high-dimensional data.
\begin{itemize}
 \item ANNs do not maintain the local relationship among the neighboring pixels in a frame.
 \item Since full connectivity is maintained throughout the network, the number of parameters are proportional to the input size.
 \end{itemize}
 
To address these problems, Lecun \etal \cite{lecun1998gradient} introduced Convolutional Neural Networks (CNN), which are also called ConvNets.
% CNNs have been applied to solve many problems related to computer vision, like object recognition \cite{lecun1998gradient,krizhevsky2012imagenet}, segmentation \cite{chen2018deeplab}, localization \cite{sermanet2013overfeat} and many more.
Extensive amount of research is being carried out on images using CNN architectures to solve many problems in computer vision and machine learning. However, their
application in video stream classification is comparatively a less explored area of research. In this paper, we performed 3D convolutions in the convolutional layers of proposed 3D CNN architecture to extract the spatial and temporal features.

\subsection{Proposed 3D CNN Model: Extracting Spatio-Temporal Features}
\label{sec:3d cnn} 
\begin{figure*}[h!]
  \centering
  \includegraphics[width=1\textwidth,height=2in]{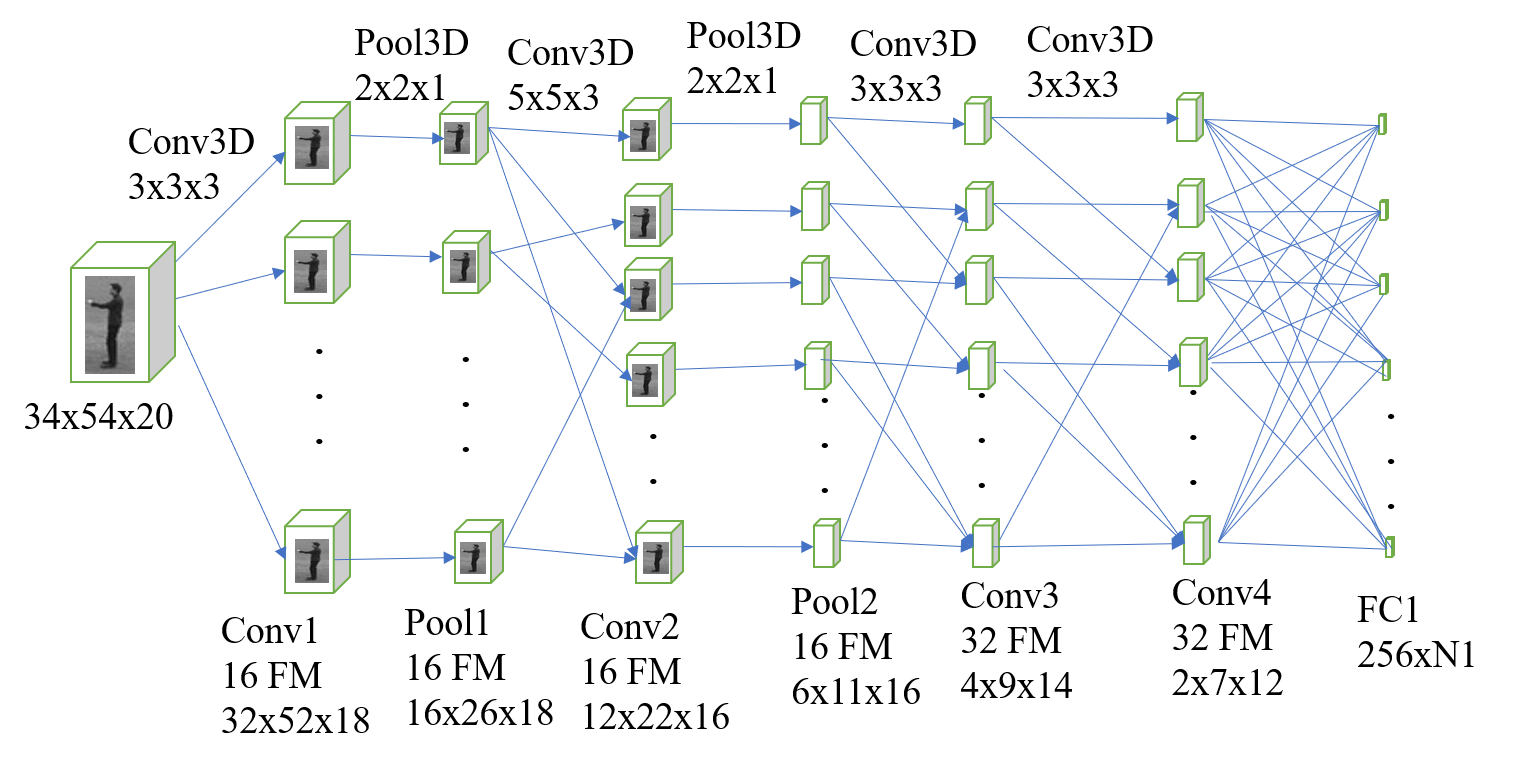}
  \caption{Proposed 3-dimensional CNN for spatio-temporal feature construction (KTH dataset). The first two convolution layers $Conv1$ and $Conv2$, both have $16$ feature maps of dimension $32\times52\times18$ and $12\times22\times16$, respectively. The $Pool1$ and $Pool2$ layers are followed by $Conv1$ and $Conv2$, to reduce the spatial dimension by half. $Conv3$ and $Conv4$ layers have 32 feature maps of dimension $4\times9\times14$ and $2\times7\times12$. Finally, a fully connected layer $FC1$ has $256$ neurons. }
  \label{fig:proposed_3D_CNN}
\end{figure*}

In 2-D CNNs, features are computed by applying the convolutions spatially over images. Whereas in case of videos, we have to consider the temporal information along with spatial features. So, it is required to extract the motion information encoded in contiguous frames using 3D convolutions. The proposed 3-dimensional CNN architecture, shown in Fig. \ref{fig:proposed_3D_CNN}, uses 3D convolutions.

Initially, the Gaussian Weighing function is used to aggregate the entire video into $20$ frames (considered 100 frames from each video throughout our experiments). To reduce the memory overhead, person centered bounding boxes are retrieved as in \cite{jhuang2007biologically,ji20133d}, which results in frames of spatial dimension $34\times 54$ and $64\times 48$ in case of KTH \cite{schuldt2004recognizing} and WEIZMANN \cite{ActionsAsSpaceTimeShapes_iccv05} datasets, respectively. 
 
In this paper, a 3D CNN model is proposed to extract spatio-temporal features, which is shown in Fig. \ref{fig:proposed_3D_CNN}. The proposed model considers the input of dimension $34\times54\times20$, corresponding to $20$ frames (encoded using GWF) of $34\times54$ pixels each. The proposed 3D CNN architecture has $5$ learnable layers, viz., $Conv1$, $Conv2$, $Conv3$, $Conv4$, and $FC1$. $Pool1$ and $Pool2$ max pooling layers are applied after $Conv1$ and $Conv2$ to reduce the spatial dimension of the feature maps by half. 
% The number of trainable layers ($Conv$, $FC$ layers) are decided based upon experimental validation on the specific datasets.
The $Conv1$ layer generates $16$ feature maps of size $32\times52\times18$ by convolving $16$ 3-D kernels of size $3\times3\times3$. $Pool1$ layer down samples the feature maps by half, after applying sub-sampling operation with a receptive field of $2\times2\times1$, which results in a $16\times26\times18$ dimensional feature vector. The $Conv2$ layer results in a $12\times22\times16$ dimensional feature map by convolving $16$ filters of size $5\times5\times3\times16$. The $Pool2$ layer produces a $6\times11\times16$ dimensional feature vector, by applying sub-sampling operation with a receptive field of $2\times2\times1$. The $3^{rd}$ convolution layer ($Conv3$) produces $32$ feature maps of dimension $4\times9\times14$, which is obtained by convolving $32$ kernels of dimension $3\times3\times3\times16$. The $Conv4$ layer generates $32$ feature maps of dimension $2\times7\times12$, which is obtained by convolving $32$ filters of dimension $3\times3\times3\times32$. The feature maps produced by $Conv4$ layer are flattened into a single feature vector of dimension $5376\times1$, which is given as input to the $1^{st}$ fully connected layer ($FC1$). Finally, the $FC1$ layer produces $256$ dimensional feature vector. The 3D CNN architecture proposed for spatio-temporal feature extraction, consists a total of $1,437,712$ trainable parameters. 
 
For WEIZMANN dataset, we used same architecture with necessary modifications. However, throughout the architecture same hyper-parameters (number of filters, filter size) are maintained as in the case of KTH dataset. The 3D CNN model proposed for WEIZMANN dataset takes input of dimension $64\times48\times20$. This model has four Conv layers ($Conv1$, $Conv2$, $Conv3$, and $Conv4$) and two max-Pooling layers ($Pool1$, $Pool2$) layers, and towards the end one fully connected layer ($FC1$). The $Conv1$ layer results in $16$ feature maps of dimension  $62\times46\times18$, which is obtained by convolving $16$ kernels of size $3\times3\times3$. The $Pool1$ layer generates reduce the spatial dimension by half, after applying sub-sampling with a receptive field of $2\times2\times1$, which generates $31\times23\times18$ dimensional feature vector. The $Conv2$ layer generates $16$ feature maps of dimension $27\times19\times16$, this is obtained by applying $16$ filters of size $5\times5\times3\times16$. The $Pool2$ layer generates a $13\times9\times16$ dimensional feature vector by sub-sampling with a receptive field of $2\times2\times1$. The $Pool2$ layer does not consider the right and bottom border feature values to avoid the dimension mismatch between input and filter size. The $Conv3$ layer results in a $11\times7\times14$ dimensional feature vector, which is obtained by convolving $32$ filters of size $3\times3\times3\times16$. The $Conv4$ layer results in $32$ feature maps of dimension $9\times5\times12$, which is obtained by convolving 32 filters of dimension $3\times3\times3\times32$. The output of $Conv4$ layer is rolled into a single column vector of dimension $17280\times1$. At the end of the architecture, $FC1$ layer has $256$ neurons, which results in a $256$ dimensional feature vector. The proposed 3D CNN architecture for WEIZMANN human action dataset consists of $4,485,136$ number of learnable parameters. The learned spatio-temporal features are given as input to LSTM model to learn the label of the entire sequence.

\subsection{Classification using Long Short-Term Memory (LSTM)}
\label{LSTM}

\begin{figure*}[h!]
  \centering
  \includegraphics[width=0.8\textwidth,height=1.2in]{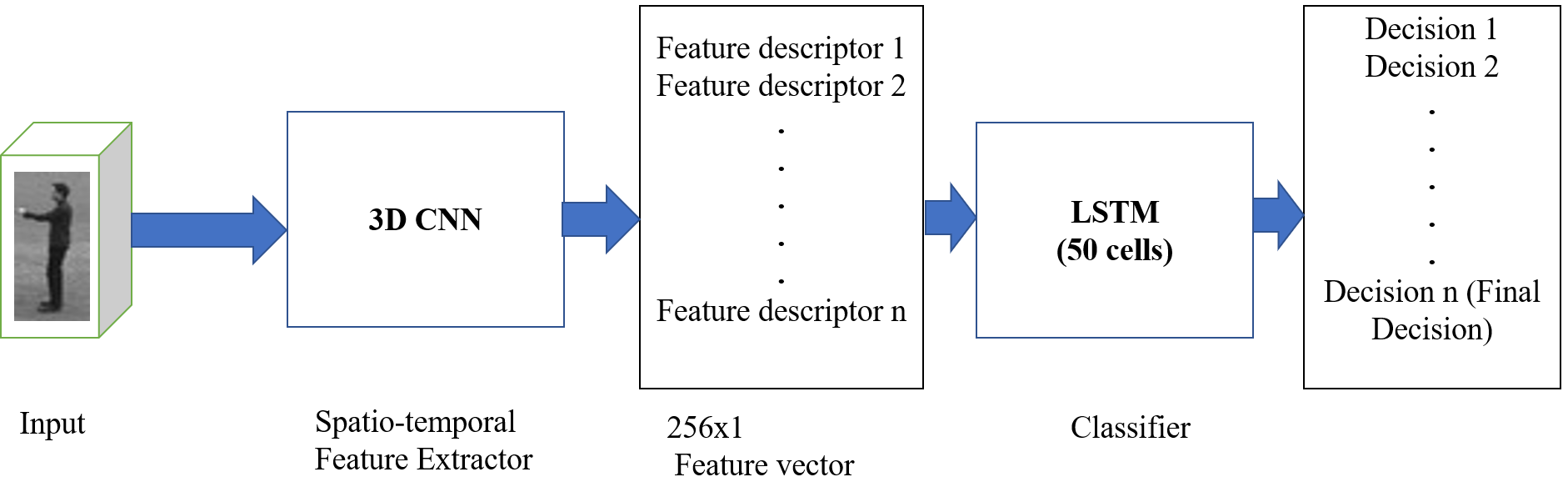}
  \caption{The proposed two-steps deep neural network approach. Encoded frames are given as input to the 3D CNN model to extract spatio-temporal features as discussed in secton \ref{sec:3d cnn} . The proposed 3D CNN model generates $256\times1$ dimensional feature vector, which is given as input the LSTM model to classify human actions. The LSTM has one hidden layer with $50$ cells, that accumulates the individual decisions corresponding to small temporal neighborhood ($4$ frames ) of the video.}
  \label{3dcnn_lstm_demo}
\end{figure*}

Once the 3D-CNN architecture is trained, it learns the spatio-temporal features automatically. The learned features are provided as input to an LSTM architecture (a Recurrent Neural Networks (RNN)) for classification. RNNs are widely used deep learning models to accumulate the individual decisions related to small temporal neighborhood of the video. RNNs make use of recurrent connections to analyze the temporal data. However, RNNs able to learn the information which are about short duration.  To learn the class label of the entire sequence, Long Short-Term Memory (LSTM) \cite{gers2002learning} is employed, which accumulates the individual decisions corresponds to each small temporal neighborhood. To obtain a sequence,  we have considered every $4$ frames as a temporal neighborhood. To classify human actions, we employ an RNN model having a hidden layer of LSTM cells. Figure \ref{3dcnn_lstm_demo} shows the overview of the proposed two-steps learning process. The input to this RNN architecture is $256$ $FC1$ features per time step. These $256$ dimensional input features are fully connected with LSTM cells. The number of LSTM cells considered are $50$ as in \cite{baccouche2011sequential}. The training details of the proposed 3D CNN architecture is presented in section \ref{training_details}.

\section{Experiments, Results and Discussions}
\label{sec:results}
As the proposed method aims to classify human actions in a video, where the videos are captured at a distance from the performer, we trained and evaluated the proposed 3D CNN model on KTH and WEIZMANN datasets. Also we experimented with transfer learning techniques, where proposed method is trained with KTH and then tested on WEIZMANN dataset, and vice versa.

\subsection{KTH dataset}
KTH dataset \cite{schuldt2004recognizing} is one among the popular datasets in human action recognition. This dataset consists of six actions, viz., walking, jogging, running, boxing, hand-waving, and hand-clapping which were carried out by $25$ persons and the videos were recorded in four different scenarios (outdoor, variations in scale, variations in cloths, and indoor).
% A few samples from KTH dataset \cite{schuldt2004recognizing} are presented in Figure \ref{kth_fig}. 
The spatial dimension of each frame is $160\times120$ pixels and the rate of frames per second (fps) is $25$. This dataset has $600$ videos. All the videos were captured from a distance from the performer. As a result, the area covered by the person is less than $10\%$ of the whole frame.

% \begin{figure*}[h!]
%   \centering
%   \includegraphics[width=0.8\textwidth]{images/kth_updated}
%   \caption{A few sample of actions from KTH dataset \cite{schuldt2004recognizing}. Six different actions are shown column-wise. The videos were recorded in four scenarios, outdoors $s1$, outdoor with scale-variation $s2$, outdoors with different cloths $s3$, and indoors $s4$, which is shown row-wise in the figure.}
%   \label{kth_fig}
% \end{figure*}

\subsection{WEIZMANN dataset}
The WEIZMANN human activity recognition dataset \cite{ActionsAsSpaceTimeShapes_iccv05} consists of $84$ videos that correspond to ten actions, which were performed by nine different people. The ten actions are gallop sideways (Side), jumping-back (jack), bending, one-hand-waving (Wave1), two-hands-waving (Wave2), walking, skipping, jumping in place (Pjump), jumping-forward (jump), and running. The spatial dimension of each frame is $180\times144$, and is at $25$ frames per second (fps). 
% A few sample of frames and corresponding action labels of WEIZMANN dataset are depicted in Fig. \ref{weizmann_fig}.
The area covered by the person is less than $12\%$ of the entire frame, due to the reason that videos were captured from a distance from the performer.

% \begin{figure*}[h!]
%   \centering
%   \includegraphics[width=0.8\textwidth]{images/weizmann_updated}
%   \caption{An illustration of actions from WEIZMANN dataset \cite{ActionsAsSpaceTimeShapes_iccv05}. Action labels are specified above the corresponding frames.} 
%   \label{weizmann_fig}
% \end{figure*}

\subsection{Experimental Results}
To validate the performance of the proposed 3-D CNN model, throughout our experiments, we have considered videos up to $4$ seconds length ($100$ frames) and aggregated them into $20$ frames using Gaussian Weighing Function as discussed in Section \ref{GWF}. To reduce the memory consumption, we have used the person-centered bounding boxes as in \cite{jhuang2007biologically,ji20133d}. Apart from these simple preprocessing steps we have not performed any other complex preprocessing like optical flow, gradients, etc. 

\subsubsection{Training Setup}
\label{training_details}
To train the proposed 3-D CNN architectures, ReLU \cite{krizhevsky2012imagenet} is used as the activation function after every $Conv$ and $FC$ layers (except output $FC$ layer). Initially learning rate is considered as $1\times10^{-4}$. The value of the learning rate reduced with a factor of $\sqrt[2]{0.1}$ after every $100$ epochs. The developed models are trained for $300$ epochs using Adam optimizer \cite{kingma2014adam} with $\beta_{1} = 0.9$, $\beta_{2} = 0.99$, and decay = $1\times10^{-6}$. The $80\%$ of entire data is used to train the 3D CNN model and remaining data is utilized to test the performance of the model. After employing Gaussian Weighting function, we obtained $20$ frames corresponding to an entire video. To reduce the amount of over-fitting, we generated $1800$ and $270$ videos (of length $20$ frames) for KTH and WEIZMANN datasets, respectively, using data-augmentation techniques like vertical flip, horizontal flip, rotation by $30^{\circ}$. We also employed dropout \cite{srivastava2014dropout} (after each $Conv$, $FC$ layers except final $FC$ layer with a rate of $0.4$, after $ReLU$ is applied) along with data augmentation to reduce the amount of over-fitting.   

\begin{table*}[!t]
\begin{center}
\caption{\label{performance_table_WEIZMANN}A performance comparison of state-of-the-art methods on WEIZMANN dataset with proposed 3D CNN model using $5$-folds cross validation test.}
\begin{threeparttable}
\begin{tabular}{| c | c | c | c | }
\hline
S.No. & Method & Features & Classification Accuracy \\
\hline
1 &
Baccouche \etal \cite{baccouche2011sequential} & 3D CNN  features & $94.58$  \\
\hline
2 & 
Gorelick \etal \cite{ActionsAsSpaceTimeShapes_iccv05} &  Space-time saliency, Action dynamics & 97.83 
\\
\hline

3
&
Fathi \etal \cite{fathi2008action} & Mid-level motion features & 100

\\
\hline
4
&
Bilen \etal \cite{bilen2016dynamic}  & 2D CNN features   & $85.2$ \\
\hline

5
&
Proposed method  & 3D CNN features   & 95.78 $\pm$ 0.58

 \\
\hline
6 & 
Proposed method applying Transfer Learning\tnote{*}  & 3D CNN features & 96.53 $\pm$ 0.07

\\
\hline
% \multicolumn{4}{@{}l}{}
\end{tabular}
  \begin{tablenotes}
    % \item[+] \small{We implemented the 3D CNN model proposed in \cite{baccouche2011sequential}, considering same sized filters and equal number of filters.}
    \item[*] \small{Fine-tuning the last two $FC$ layers of pre-trained model, which is trained on KTH dataset.}
  \end{tablenotes}
\end{threeparttable}
\end{center}
\end{table*}

\subsubsection{Results and Discussions}
The obtained results are compared with the state-of-the-art methods as shown in Tables \ref{performance_table_WEIZMANN}, \ref{performance_table_KTH} on WEIZMANN  and KTH datasets, respectively. \cite{baccouche2011sequential} reported $94.39 \%$ accuracy over KTH dataset using a 3D CNN architecture having five trainable layers. However, they have not evaluated their model on WEIZMANN dataset, we obtained $94.58\%$ accuracy through our experiment (input dimension is 64x48x9) using the same architecture (same w.r.t number of features, filter size, number of neurons in $FC$ layers) as in \cite{baccouche2011sequential}. 
After employing the proposed scheme of generating aggregated video to the 3D CNN model proposed in \cite{baccouche2011sequential}, we observed that the model outperforming the original model. However, the Dynamic Image Network introduced by Bilen \etal \cite{bilen2016dynamic} results in high amount of over-fitting due to which it produces only $85.2\%$, $86.8\%$ accuracies for KTH, WEIZMANN datasets. The proposed 3D CNN model produces $95.27\%$ and $95.78\%$ accuracies on KTH and WEIZMANN datasets, respectively, when the size of Gaussian weight vector is $5$. From Table \ref{performance_table_WEIZMANN} and Table \ref{performance_table_KTH}, we can observe that the proposed 3D CNN model outperforming other deep learning based models on both the datasets.    

\begin{table*}[!t]
\caption{\label{performance_table_KTH}Comparing the state-of-the-art human action recognition approaches on KTH dataset with the proposed 3D CNN model using $5$-folds cross validation test.}
\centering
\begin{threeparttable}
\begin{tabular}{|p{1cm}|p{4.5cm}|l|p{3.5cm}|}
\hline
S.No. & Method & Features & Classification Accuracy \\
\hline
1 &
 Nazir \etal \cite{nazir2018bag} & Bag of Expressions (BoE)  & 99.51  \\ \hline
2 & 
Baccouche \etal \cite{baccouche2011sequential} & 3D CNN  features   & 94.39   
\\
\hline

3
&
 Abdulmunem \etal \cite{abdulmunem2016saliency}  & Bag of Visual Words & 97.20
\\
\hline
4
&
Ji \etal \cite{ji20133d}  & 3D CNN features & 90.20
 \\
\hline
5 & 
  Wang \etal \cite{wang2013dense}  & Dense Trajectories and motion boundary descriptor & 95.00
\\ \hline

6 &   Gilbert \etal \cite{gilbert2011action}  & Mined Hierarchical compound features & 94.50 \\ \hline

7 & Yang \etal \cite{yang2015action}  & Multi-scale oriented neighborhood features & 96.50 \\ \hline

8 & Kovashka \etal \cite{kovashka2010learning}  & Hierarchical Space time neighborhood features & 94.53 \\
\hline
9 & Bilen \etal \cite{bilen2016dynamic}  & 2D CNN features & $86.8$ \\
\hline

10 &  Baccouche \etal \cite{baccouche2011sequential}\tnote{\#} & 3D CNN  features  & 94.78 $\pm$ 0.11 \\ \hline
11 & Proposed Method  & 3D CNN  features  & 95.27 $\pm$ 0.45 \\ \hline

12 &  Proposed Method applying Transfer learning\tnote{\$}  & 3D CNN  features & 95.86 $\pm$ 0.3 \\ \hline

% \multicolumn{4}{@{}l}{}
\end{tabular}
  \begin{tablenotes}
    \item[\#] \small{Encoded frames are given as input to the 3D CNN model proposed in \cite{baccouche2011sequential}. (the size of Gaussian vector is $5$)}. 
    \item[\$] \small{Fine-tuning the last two $FC$ layers of pre-trained model, which is trained on WEIZMANN dataset.}
  \end{tablenotes}
\end{threeparttable}
\end{table*}

\begin{figure}[h!]
  \centering
  \includegraphics[width=0.5 \textwidth,height=1.25in]{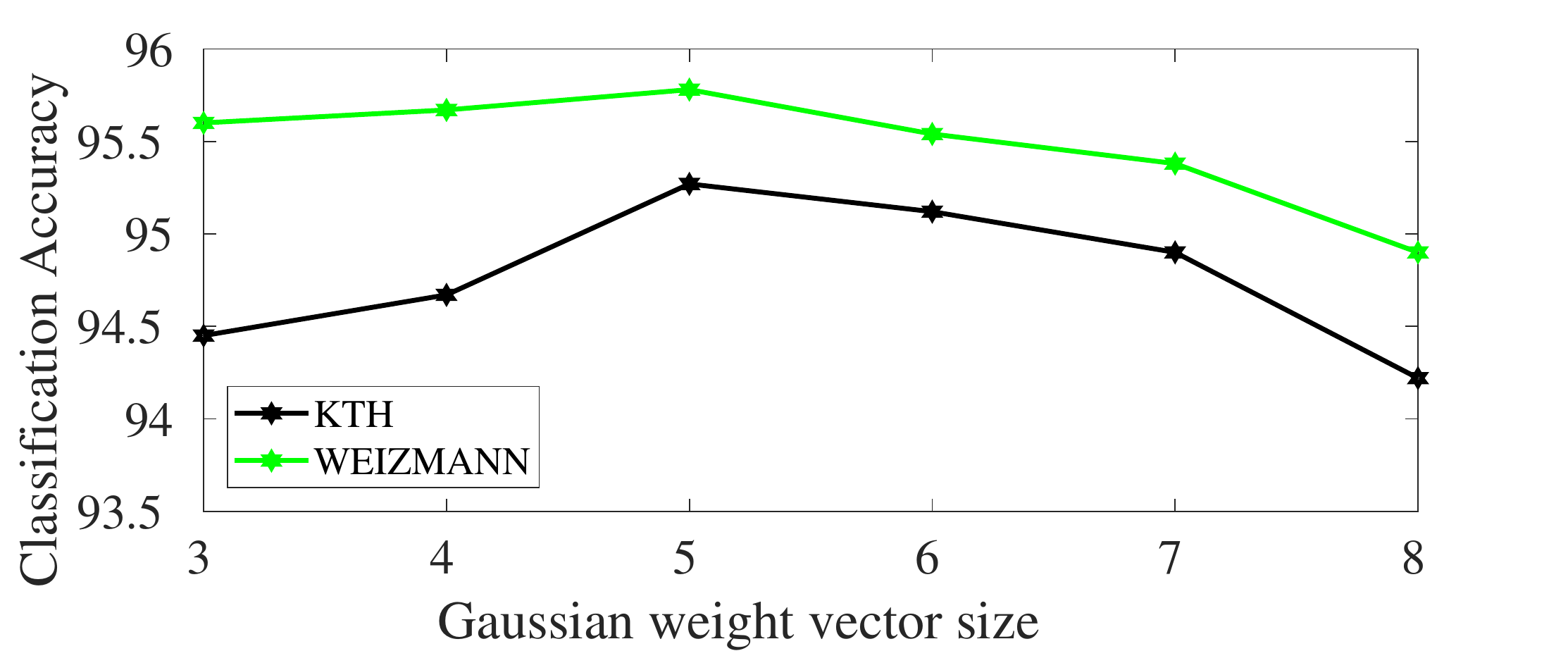}
  \caption{A performance comparison of proposed 3D CNN model by varying the size of Gaussian weight vector. The size of the Gaussian weight vector is considered as $3$, $4$, $5$, $6$, $7$, and $8$ in our experiments.}
  \label{GWF_g}
\end{figure}

When compared with human action recognition methods involving hand-crafted features, our method produces comparable results with state-of-the-art on both KTH and WEIZMANN datasets. We also experimented the performance of our model by varying the size of Gaussian weight vector $W$ in the range from $3$ to $8$. The performance variation of proposed model is shown in Figure \ref{GWF_g} by varying the size of Gaussian weight vector $W^\prime$. We observe that the proposed 3D CNN architecture is showing the best accuracy, when the size of Gaussian weight vector $W$ = $5$. Based on the results depicted in Table \ref{performance_table_WEIZMANN} and Table \ref{performance_table_KTH}, we can conclude that, our 3D CNN architecture outperforms the state-of-the-art deep learning architectures.
% when applied to the datasets containing videos captured by camera placed far from the performer. 
However, due to the small size of the available dataset of such kind, the proposed deep learning based method could not outperform the hand-crafted feature based methods (although showing a comparable result).

\begin{figure}[h!]
  \centering
  \includegraphics[width=0.5 \textwidth, height=1.25in]{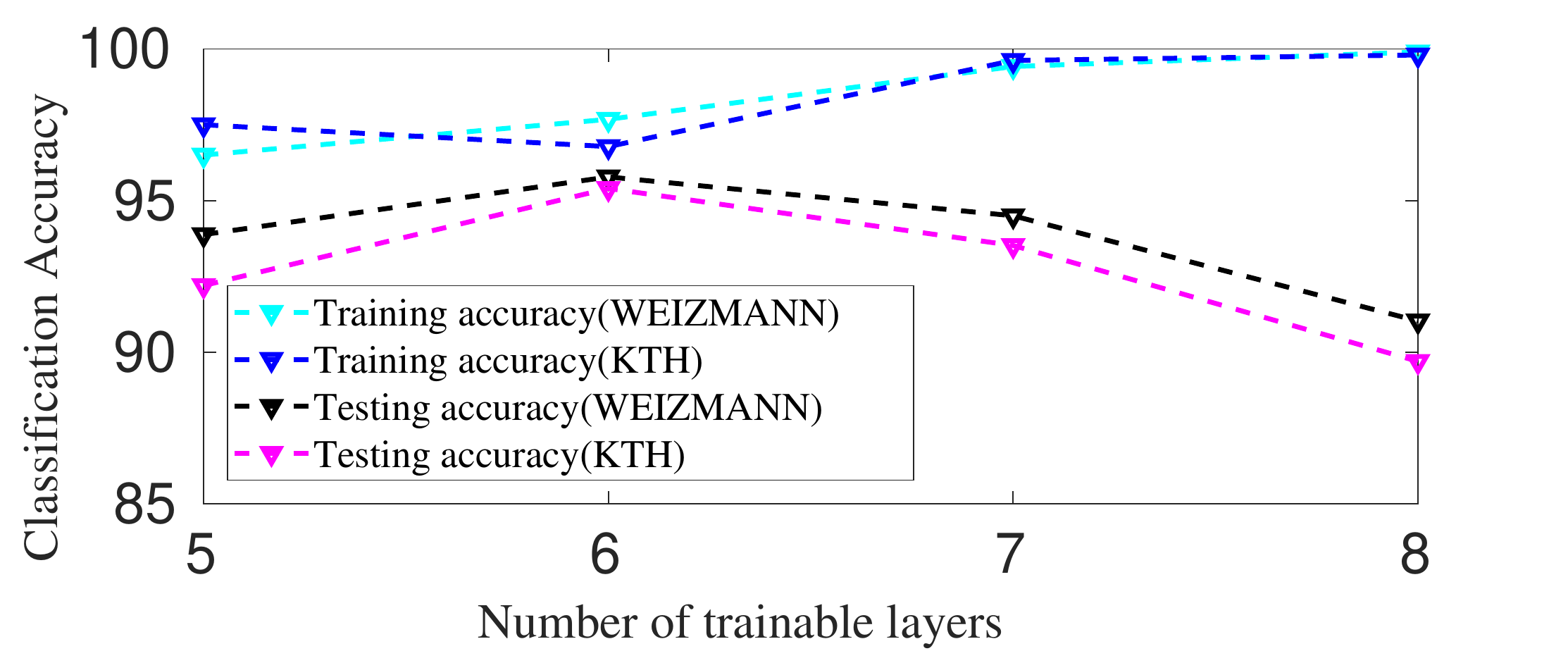}
  \caption{Comparing the Training and Testing accuracies of both the datasets  by varying the number of trainable layers ($5$, $6$, $7$, and $8$) in the proposed 3D CNN architecture.}
  \label{performance_layer1}
\end{figure}

Basha \etal \cite{basha2020impact} shown the necessity of the fully connected layers based on the depth of the CNN. Motivated by their work, experiments are conducted by varying the number of trainable layers in the proposed 3D CNN architecture. The amount of over-fitting increases in the context of both the datasets after inclusion of more $FC$ layers. 
% At the same time, whenever the flattened features are directly given to the soft-max layer the proposed model exhibits lower performance. 
 The performance of the proposed 3D CNN architecture with varying number of trainable layers is depicted in Figure \ref{performance_layer1}.  

A common practice in deep learning community (especially to deal with small datasets) is that, using the pre-trained models to reduce the training time and obtaining competitive results by training the models for a fewer number of epochs. Generally, these pre-trained models work as feature extractors. With this motivation, we utilized the pre-trained model of KTH dataset to fine-tune over WEIZMANN dataset and vice-versa. The last two layers ($Conv4$, $FC1$) of the proposed 3D CNN model are fine-tuned in both the cases. Results of the above experiments are reported in the last rows of the Table \ref{performance_table_WEIZMANN} and Table \ref{performance_table_KTH}, respectively. We can observe a little increase in the classification accuracy for both the datasets, after applying the above scheme.

\section{Conclusion}
\label{sec:conclusion}
We introduced an information-rich sampling technique using Gaussian weighing function as a pre-processing step before giving it as input to any deep learning model, for better classification of human actions from videos. The proposed scheme aggregates consecutive $k$ frames into a single frame by applying a Gaussian weighted summation of the $k$ frames. We further proposed a 3D CNN model that learns and extracts spatio-temporal features by performing 3D convolutions. The classification of the human actions are performed using LSTM. %The proposed 3D CNN model over LSTM, can deal with videos captured from camera placed at a distance from the performer. 
Experimental results on both KTH and WEIZMANN datasets show that proposed model produces comparable results, among the state-of-the art. Whereas, the proposed 3D CNN model outperforms the state-of-the-art deep CNN models. Learning the weights for frame aggregation may be a potential future research direction. 
% Due to the unavailability of large benchmark datasets where the performers look small in the video, the proposed method could not be tested on large datasets. 
% In future, a large dataset with such videos may be proposed to test the proposed architecture and compare with the state-of-the-art methods.
% \addtolength{\parskip}{-2.5mm}

\section*{Acknowledgments}
We acknowledge the support of NVIDIA with the donation of the GeForce Titan XP GPU used for this research.
\bibliographystyle{IEEEtran}
\bibliography{refs}

% Generated by IEEEtran.bst, version: 1.14 (2015/08/26)
\begin{thebibliography}{10}
\providecommand{\url}[1]{#1}
\csname url@samestyle\endcsname
\providecommand{\newblock}{\relax}
\providecommand{\bibinfo}[2]{#2}
\providecommand{\BIBentrySTDinterwordspacing}{\spaceskip=0pt\relax}
\providecommand{\BIBentryALTinterwordstretchfactor}{4}
\providecommand{\BIBentryALTinterwordspacing}{\spaceskip=\fontdimen2\font plus
\BIBentryALTinterwordstretchfactor\fontdimen3\font minus
  \fontdimen4\font\relax}
\providecommand{\BIBforeignlanguage}[2]{{%
\expandafter\ifx\csname l@#1\endcsname\relax
\typeout{** WARNING: IEEEtran.bst: No hyphenation pattern has been}%
\typeout{** loaded for the language `#1'. Using the pattern for}%
\typeout{** the default language instead.}%
\else
\language=\csname l@#1\endcsname
\fi
#2}}
\providecommand{\BIBdecl}{\relax}
\BIBdecl

\bibitem{ziaeefard2015semantic}
M.~Ziaeefard and R.~Bergevin, ``Semantic human activity recognition: A
  literature review,'' \emph{Pattern Recognition}, vol.~48, no.~8, pp.
  2329--2345, 2015.

\bibitem{laptev2007retrieving}
I.~Laptev and P.~P{\'e}rez, ``Retrieving actions in movies,'' in \emph{Computer
  Vision, 2007. ICCV 2007. IEEE 11th International Conference on}.\hskip 1em
  plus 0.5em minus 0.4em\relax IEEE, 2007, pp. 1--8.

\bibitem{dollar2005behavior}
P.~Doll{\'a}r, V.~Rabaud, G.~Cottrell, and S.~Belongie, ``Behavior recognition
  via sparse spatio-temporal features,'' in \emph{Visual Surveillance and
  Performance Evaluation of Tracking and Surveillance, 2005. 2nd Joint IEEE
  International Workshop on}.\hskip 1em plus 0.5em minus 0.4em\relax IEEE,
  2005, pp. 65--72.

\bibitem{Sun_2017_ICCV}
L.~Sun, K.~Jia, K.~Chen, D.-Y. Yeung, B.~E. Shi, and S.~Savarese, ``Lattice
  long short-term memory for human action recognition,'' in \emph{The IEEE
  International Conference on Computer Vision (ICCV)}, Oct 2017.

\bibitem{simonyan2014two}
K.~Simonyan and A.~Zisserman, ``Two-stream convolutional networks for action
  recognition in videos,'' in \emph{Advances in neural information processing
  systems}, 2014, pp. 568--576.

\bibitem{wang2016temporal}
L.~Wang, Y.~Xiong, Z.~Wang, Y.~Qiao, D.~Lin, X.~Tang, and L.~Van~Gool,
  ``Temporal segment networks: Towards good practices for deep action
  recognition,'' in \emph{European Conference on Computer Vision}.\hskip 1em
  plus 0.5em minus 0.4em\relax Springer, 2016, pp. 20--36.

\bibitem{baccouche2011sequential}
M.~Baccouche, F.~Mamalet, C.~Wolf, C.~Garcia, and A.~Baskurt, ``Sequential deep
  learning for human action recognition,'' in \emph{International Workshop on
  Human Behavior Understanding}.\hskip 1em plus 0.5em minus 0.4em\relax
  Springer, 2011, pp. 29--39.

\bibitem{nazir2018bag}
S.~Nazir, M.~H. Yousaf, J.-C. Nebel, and S.~A. Velastin, ``A bag of expression
  framework for improved human action recognition,'' \emph{Pattern Recognition
  Letters}, 2018.

\bibitem{malgireddy2013language}
M.~R. Malgireddy, I.~Nwogu, and V.~Govindaraju, ``Language-motivated approaches
  to action recognition,'' \emph{The Journal of Machine Learning Research},
  vol.~14, no.~1, pp. 2189--2212, 2013.

\bibitem{chaudhry2009histograms}
R.~Chaudhry, A.~Ravichandran, G.~Hager, and R.~Vidal, ``Histograms of oriented
  optical flow and binet-cauchy kernels on nonlinear dynamical systems for the
  recognition of human actions,'' in \emph{2009 IEEE Conference on Computer
  Vision and Pattern Recognition}.\hskip 1em plus 0.5em minus 0.4em\relax IEEE,
  2009, pp. 1932--1939.

\bibitem{mukherjee2015human}
S.~Mukherjee, ``Human action recognition using dominant pose duplet,'' in
  \emph{International Conference on Computer Vision Systems}.\hskip 1em plus
  0.5em minus 0.4em\relax Springer, 2015, pp. 488--497.

\bibitem{mukherjee2}
S.~Mukherjee, S.~K. Biswas, and D.~P. Mukherjee, ``Recognizing interactions
  between human performers by ‘dominating pose doublet’,'' \emph{Machine
  Vision and Applications}, vol.~25, no.~4, pp. 1033--1052, 2014.

\bibitem{wang2013dense}
H.~Wang, A.~Kl{\"a}ser, C.~Schmid, and C.-L. Liu, ``Dense trajectories and
  motion boundary descriptors for action recognition,'' \emph{International
  journal of computer vision}, vol. 103, no.~1, pp. 60--79, 2013.

\bibitem{laptev2005space}
I.~Laptev, ``On space-time interest points,'' \emph{International journal of
  computer vision}, vol.~64, no. 2-3, pp. 107--123, 2005.

\bibitem{harris1988combined}
C.~G. Harris, M.~Stephens \emph{et~al.}, ``A combined corner and edge
  detector.'' in \emph{Alvey vision conference}, vol.~15, no.~50.\hskip 1em
  plus 0.5em minus 0.4em\relax Citeseer, 1988, pp. 10--5244.

\bibitem{chen2009mosift}
M.-y. Chen and A.~Hauptmann, ``Mosift: Recognizing human actions in
  surveillance videos,'' 2009.

\bibitem{dawn2016comprehensive}
D.~D. Dawn and S.~H. Shaikh, ``A comprehensive survey of human action
  recognition with spatio-temporal interest point (stip) detector,'' \emph{The
  Visual Computer}, vol.~32, no.~3, pp. 289--306, 2016.

\bibitem{buddubariki2016event}
V.~Buddubariki, S.~G. Tulluri, and S.~Mukherjee, ``Event recognition in
  egocentric videos using a novel trajectory based feature,'' in
  \emph{Proceedings of the Tenth Indian Conference on Computer Vision, Graphics
  and Image Processing}.\hskip 1em plus 0.5em minus 0.4em\relax ACM, 2016,
  p.~76.

\bibitem{abdulmunem2016saliency}
A.~Abdulmunem, Y.-K. Lai, and X.~Sun, ``Saliency guided local and global
  descriptors for effective action recognition,'' \emph{Computational Visual
  Media}, vol.~2, no.~1, pp. 97--106, 2016.

\bibitem{ji20133d}
S.~Ji, W.~Xu, M.~Yang, and K.~Yu, ``3d convolutional neural networks for human
  action recognition,'' \emph{IEEE transactions on pattern analysis and machine
  intelligence}, vol.~35, no.~1, pp. 221--231, 2013.

\bibitem{taylor2010convolutional}
G.~W. Taylor, R.~Fergus, Y.~LeCun, and C.~Bregler, ``Convolutional learning of
  spatio-temporal features,'' in \emph{European conference on computer
  vision}.\hskip 1em plus 0.5em minus 0.4em\relax Springer, 2010, pp. 140--153.

\bibitem{tran2015learning}
D.~Tran, L.~Bourdev, R.~Fergus, L.~Torresani, and M.~Paluri, ``Learning
  spatiotemporal features with 3d convolutional networks,'' in \emph{Computer
  Vision (ICCV), 2015 IEEE International Conference on}.\hskip 1em plus 0.5em
  minus 0.4em\relax IEEE, 2015, pp. 4489--4497.

\bibitem{karpathy2014large}
A.~Karpathy, G.~Toderici, S.~Shetty, T.~Leung, R.~Sukthankar, and L.~Fei-Fei,
  ``Large-scale video classification with convolutional neural networks,'' in
  \emph{Proceedings of the IEEE conference on Computer Vision and Pattern
  Recognition}, 2014, pp. 1725--1732.

\bibitem{kar2017adascan}
A.~Kar, N.~Rai, K.~Sikka, and G.~Sharma, ``Adascan: Adaptive scan pooling in
  deep convolutional neural networks for human action recognition in videos,''
  in \emph{Proceedings of the IEEE Conference on Computer Vision and Pattern
  Recognition}, 2017, pp. 3376--3385.

\bibitem{herath2017going}
S.~Herath, M.~Harandi, and F.~Porikli, ``Going deeper into action recognition:
  A survey,'' \emph{Image and vision computing}, vol.~60, pp. 4--21, 2017.

\bibitem{srivastava2015unsupervised}
N.~Srivastava, E.~Mansimov, and R.~Salakhudinov, ``Unsupervised learning of
  video representations using lstms,'' in \emph{International conference on
  machine learning}, 2015, pp. 843--852.

\bibitem{bilen2016dynamic}
H.~Bilen, B.~Fernando, E.~Gavves, A.~Vedaldi, and S.~Gould, ``Dynamic image
  networks for action recognition,'' in \emph{Proceedings of the IEEE
  Conference on Computer Vision and Pattern Recognition}, 2016, pp. 3034--3042.

\bibitem{schuldt2004recognizing}
C.~Schuldt, I.~Laptev, and B.~Caputo, ``Recognizing human actions: a local svm
  approach,'' in \emph{Pattern Recognition, 2004. ICPR 2004. Proceedings of the
  17th International Conference on}, vol.~3.\hskip 1em plus 0.5em minus
  0.4em\relax IEEE, 2004, pp. 32--36.

\bibitem{gers2002learning}
F.~A. Gers, N.~N. Schraudolph, and J.~Schmidhuber, ``Learning precise timing
  with lstm recurrent networks,'' \emph{Journal of machine learning research},
  vol.~3, no. Aug, pp. 115--143, 2002.

\bibitem{lecun1998gradient}
Y.~LeCun, L.~Bottou, Y.~Bengio, and P.~Haffner, ``Gradient-based learning
  applied to document recognition,'' \emph{Proceedings of the IEEE}, vol.~86,
  no.~11, pp. 2278--2324, 1998.

\bibitem{jhuang2007biologically}
H.~Jhuang, T.~Serre, L.~Wolf, and T.~Poggio, ``A biologically inspired system
  for action recognition,'' in \emph{Computer Vision, 2007. ICCV 2007. IEEE
  11th International Conference on}.\hskip 1em plus 0.5em minus 0.4em\relax
  Ieee, 2007, pp. 1--8.

\bibitem{ActionsAsSpaceTimeShapes_iccv05}
L.~Gorelick, M.~Blank, E.~Shechtman, M.~Irani, and R.~Basri, ``Actions as
  space-time shapes,'' \emph{IEEE transactions on pattern analysis and machine
  intelligence}, vol.~29, no.~12, pp. 2247--2253, 2007.

\bibitem{krizhevsky2012imagenet}
A.~Krizhevsky, I.~Sutskever, and G.~E. Hinton, ``Imagenet classification with
  deep convolutional neural networks,'' in \emph{Advances in neural information
  processing systems}, 2012, pp. 1097--1105.

\bibitem{kingma2014adam}
D.~P. Kingma and J.~Ba, ``Adam: A method for stochastic optimization,''
  \emph{arXiv preprint arXiv:1412.6980}, 2014.

\bibitem{srivastava2014dropout}
N.~Srivastava, G.~Hinton, A.~Krizhevsky, I.~Sutskever, and R.~Salakhutdinov,
  ``Dropout: A simple way to prevent neural networks from overfitting,''
  \emph{The Journal of Machine Learning Research}, vol.~15, no.~1, pp.
  1929--1958, 2014.

\bibitem{fathi2008action}
A.~Fathi and G.~Mori, ``Action recognition by learning mid-level motion
  features,'' in \emph{Computer Vision and Pattern Recognition, 2008. CVPR
  2008. IEEE Conference on}.\hskip 1em plus 0.5em minus 0.4em\relax IEEE, 2008,
  pp. 1--8.

\bibitem{gilbert2011action}
A.~Gilbert, J.~Illingworth, and R.~Bowden, ``Action recognition using mined
  hierarchical compound features,'' \emph{IEEE Transactions on Pattern Analysis
  and Machine Intelligence}, vol.~33, no.~5, pp. 883--897, 2011.

\bibitem{yang2015action}
J.~Yang, Z.~Ma, and M.~Xie, ``Action recognition based on multi-scale oriented
  neighborhood features,'' \emph{International Journal of Signal Processing,
  Image Processing and Pattern Recognition}, vol.~8, no.~1, pp. 241--254, 2015.

\bibitem{kovashka2010learning}
A.~Kovashka and K.~Grauman, ``Learning a hierarchy of discriminative space-time
  neighborhood features for human action recognition,'' in \emph{Computer
  Vision and Pattern Recognition (CVPR), 2010 IEEE Conference on}.\hskip 1em
  plus 0.5em minus 0.4em\relax IEEE, 2010, pp. 2046--2053.

\bibitem{basha2020impact}
S.~S. Basha, S.~R. Dubey, V.~Pulabaigari, and S.~Mukherjee, ``Impact of fully
  connected layers on performance of convolutional neural networks for image
  classification,'' \emph{Neurocomputing}, vol. 378, pp. 112--119, 2020.

\end{thebibliography}

% \section*{Supplementary Material}

% Supplementary material that may be helpful in the review process should
% be prepared and provided as a separate electronic file. That file can
% then be transformed into PDF format and submitted along with the
% manuscript and graphic files to the appropriate editorial office.
\end{document}